\documentclass[fleqn,final,5p,times]{elsarticle}

\usepackage{amssymb}
\usepackage{amsthm}
\usepackage{amsmath}

\usepackage{algpseudocode}
\usepackage{algorithm}

\usepackage[hidelinks]{hyperref}

\newcommand{\figref}[1]{Figure~\ref{#1}}
\newcommand{\tabref}[1]{Table~\ref{#1}}
\newcommand{\secref}[1]{Section~\ref{#1}}
\newcommand{\algoref}[1]{Algorithm~\ref{#1}}
\LetLtxMacro{\originaleqref}{\eqref}
\renewcommand{\eqref}{Eq.~\originaleqref}

\begin{document}

\begin{frontmatter}

\title{Deep Reinforcement Learning for Computational Fluid Dynamics on HPC Systems}

\author[label1]{Marius Kurz\fnref{fn1}\corref{cor1}}
\ead{marius.kurz@iag.uni-stuttgart.de}

\author[label2]{Philipp Offenh\"auser\fnref{fn1}}
\ead{philipp.offenhaeuser@hpe.com}

\author[label2]{Dominic Viola}
\ead{dominic.viola@hpe.com}

\author[label3]{Oleksandr Shcherbakov}
\ead{oleksandr.shcherbakov@hlrs.com}

\author[label3]{Michael Resch}
\ead{resch@hlrs.de}

\author[label4]{Andrea Beck}
\ead{beck@iag.uni-stuttgart.de}

\fntext[fn1]{M. Kurz and P. Offenh\"auser share first authorship.}
\address[label1]{Institute of Aerodynamics and Gas Dynamics, University of Stuttgart, Pfaffenwaldring 21, 70569 Stuttgart, Germany}
\address[label2]{Hewlett Packard Enterprise (HPE), Herrenberger Straße 140, 71034  Böblingen, Germany}
\address[label3]{High Performance Computing Center Stuttgart (HLRS), University of Stuttgart, Nobelstraße 19, 70569 Stuttgart, Germany}
\address[label4]{Laboratory of Fluid Dynamics and Technical Flows, University of Magdeburg ``Otto von Guericke'', Universitätsplatz 2, 39106 Madgeburg, Germany}

\cortext[cor1]{Corresponding author}

\begin{abstract}
  Reinforcement learning (RL) is highly suitable for devising control strategies in the context of dynamical systems. %
  A prominent instance of such a dynamical system is the system of equations governing fluid dynamics.
  Recent research results indicate that RL-augmented computational fluid dynamics (CFD) solvers can exceed the current state of the art, for example in the field of turbulence modeling.
  However, while in supervised learning, the training data can be generated a priori in an offline manner, RL requires constant run-time interaction and data exchange with the CFD solver during training.
  In order to leverage the potential of RL-enhanced CFD, the interaction between the CFD solver and the RL algorithm thus have to be implemented efficiently on high-performance computing (HPC) hardware. %
  To this end, we present Relexi as a scalable RL framework that bridges the gap between machine learning workflows and modern CFD solvers on HPC systems providing both components with its specialized hardware.
  Relexi is built with modularity in mind and allows easy integration of various HPC solvers by means of the in-memory data transfer provided by the SmartSim library.
  Here, we demonstrate that the Relexi framework can scale up to hundreds of parallel environment on thousands of cores.
  This allows to leverage modern HPC resources to either enable larger problems or faster turnaround times.
  Finally, we demonstrate the potential of an RL-augmented CFD solver by finding a control strategy for optimal eddy viscosity selection in large eddy simulations.  
\end{abstract}

\begin{keyword}
  Deep Reinforcement Learning \sep High-Performance Computing \sep Computational Fluid Dynamics \sep Turbulence Modeling \sep Large Eddy Simulation
\end{keyword}

\end{frontmatter}

\section{Introduction}
In recent years, there have been increasing efforts to transfer advances in machine learning (ML) to the field of computational fluid dynamics (CFD) in order to enhance simulations for a myriad of different applications \cite{brunton2020machine}, which cover the fields of turbulence modeling \cite{beck2021perspective,kurz2022machine,beck2019deep,maulik2019subgrid}, shock detection \cite{ray2019detecting,beck2020neural}, the formulation of turbulent inflow conditions \cite{fukami2019synthetic,kim2020deep} or applications in flow control \cite{paris2021robust}.
As of today, most of these advances are based on the supervised learning (SL) paradigm.
In SL, the ML model is trained based on a dataset that is obtained a priori from CFD simulations or experiments.
During training, the parameters of the ML model are optimized to approximate the functional relationship between the input and output quantities of the dataset.
Since the training dataset is fixed and does not interact with the predictions of the ML model, the training itself is thus, in a sense, static and offline.
This oftentimes leads to inconsistencies, if, during inference, the SL-trained model is confronted with a varying and dynamic environment.
In this case, the model's prior predictions affect how the system evolves and thus which input states the model will see in the future.
Ensuring that all potential states of a non-linear dynamical system are sufficiently represented within the training dataset is by no means trivial and for many applications elusive.
The reinforcement learning (RL) paradigm addresses this issue by training ML models not on a static dataset, but instead trains models by letting the model interact with the actual environment it will later be deployed in.
Thus, the training process not only aims at predicting outputs, but it does so by taking dynamically generated inputs into account.
The goal of RL is to find an optimal strategy for moving forward from the current, observed state.
In the course of the training process, predictions of the model change the state of the environment and the model will be rewarded based on how beneficial this transition is.
Training based on such sequences of actions and transitions allows incorporation of the long-term implications of the model's predictions into the training process.
Thus, the training itself becomes dynamic and requires ``online'' joint runs of the environment and the model.

CFD simulations are typically computationally expensive and thus rely heavily on high-performance computing (HPC) systems.
Coupling such HPC flow solvers with the more recently emerged ML libraries is tough, since both rely on different hardware, algorithms and overall programming paradigms.
For instance, most HPC codes are still written and optimized for CPU architectures and are parallelized with the Message Passing Interface (MPI), while ML libraries run most efficiently on GPUs.
The HPC environment thus has to provide sufficient hardware resources for both components of the application and efficient communication between them.
In addition, most HPC codes are written in compiled languages like C/C++ or Fortran.
On the other hand, ML libraries are oftentimes also written in compiled languages but are exposed to the user via a Python interface.
As these languages typically lack a well-defined interface, bridging the language gap in an efficient manner for HPC is tough.
These problems are less pronounced for SL tasks, since the generation of the training dataset with HPC codes and the training itself are mostly decoupled and can be performed separately on appropriate hardware.
The HPC solver and the ML model only have to be coupled for inference, i.e. when the trained ML model is evaluated in actual simulations.
For this, Maulik et al. \cite{maulik2021deploying} proposed to link the TensorFlow library directly to the flow solver and execute the model via TensorFlow's C-API.
Ott et al. \cite{ott2020fortran} proposed the Fortran-Keras-Bridge, which allows convenient execution of trained Keras models from Fortran code, and is based on the Neural-Fortran library by Curcic \cite{curcic2019parallel}.
For RL however, the integration of ML and HPC workloads is much more involved, since the training process itself requires running simulations and thus needs to run HPC simulations and the optimization of the ML model in parallel.
An efficient RL framework for HPC workloads needs to manage the simulations in the HPC environment and has to implement efficient communication between the simulation codes and the employed ML library.
In \cite{novati2021automating}, Novati et al. used the smarties library to couple a flow solver and TensorFlow in order to train a data-driven turbulence model for large eddy simulation (LES) on an HPC system.
Bae and Koumoutsakos applied a similar framework to the wall-modeling of wall-bounded flows in \cite{bae2021scientific}.
Pawar and Maulik \cite{pawar2021distributed} developed a distributed RL framework called PAR-RL, which they applied to optimize the time-stepping of a CFD simulation.
Rabault and colleagues developed an RL framework for flow control \cite{rabault2019accelerating,rabault2019artificial,tang2020robust}, which was also coupled with a spectral flow solver by Li and Zhang in \cite{li2022reinforcement}.
Similarly, Fan et al. \cite{fan2020reinforcement} coupled a spectral element LES solver with TensorFlow to control the flow around a cylinder.

However, most of those applications are limited in the problem sizes that can be investigated, since the simulations cannot take advantage of the parallel computing resources provided by modern HPC systems.
In this paper, we present a scalable RL framework that overcomes the gap between numerical simulation and ML workflows on HPC systems providing both components with its required specialized hardware.
Moreover, we demonstrate the prospects of the reinforcement learning paradigm in scientific computing by applying our framework to derive data-driven turbulence models for large eddy simulation. 

The paper is structured as follows.
\secref{sec:rl} gives a brief outline of the reinforcement learning paradigm.
In \secref{sec:software}, we present the different software components of our framework and how they are integrated.
The hardware environment for our benchmarks is presented in \secref{sec:hardware}.
In \secref{sec:numerics}, we outline how our framework can be applied to derive turbulence models for large eddy simulations.
The results of our scaling studies and the performance of the derived data-driven turbulence models are presented in \secref{sec:results}.
\secref{sec:conclusion} summarizes and concludes the paper.

\section{Reinforcement Learning - A Brief Outline}\label{sec:rl}

\begin{figure}
  \centering
  \includegraphics[width=0.9\linewidth]{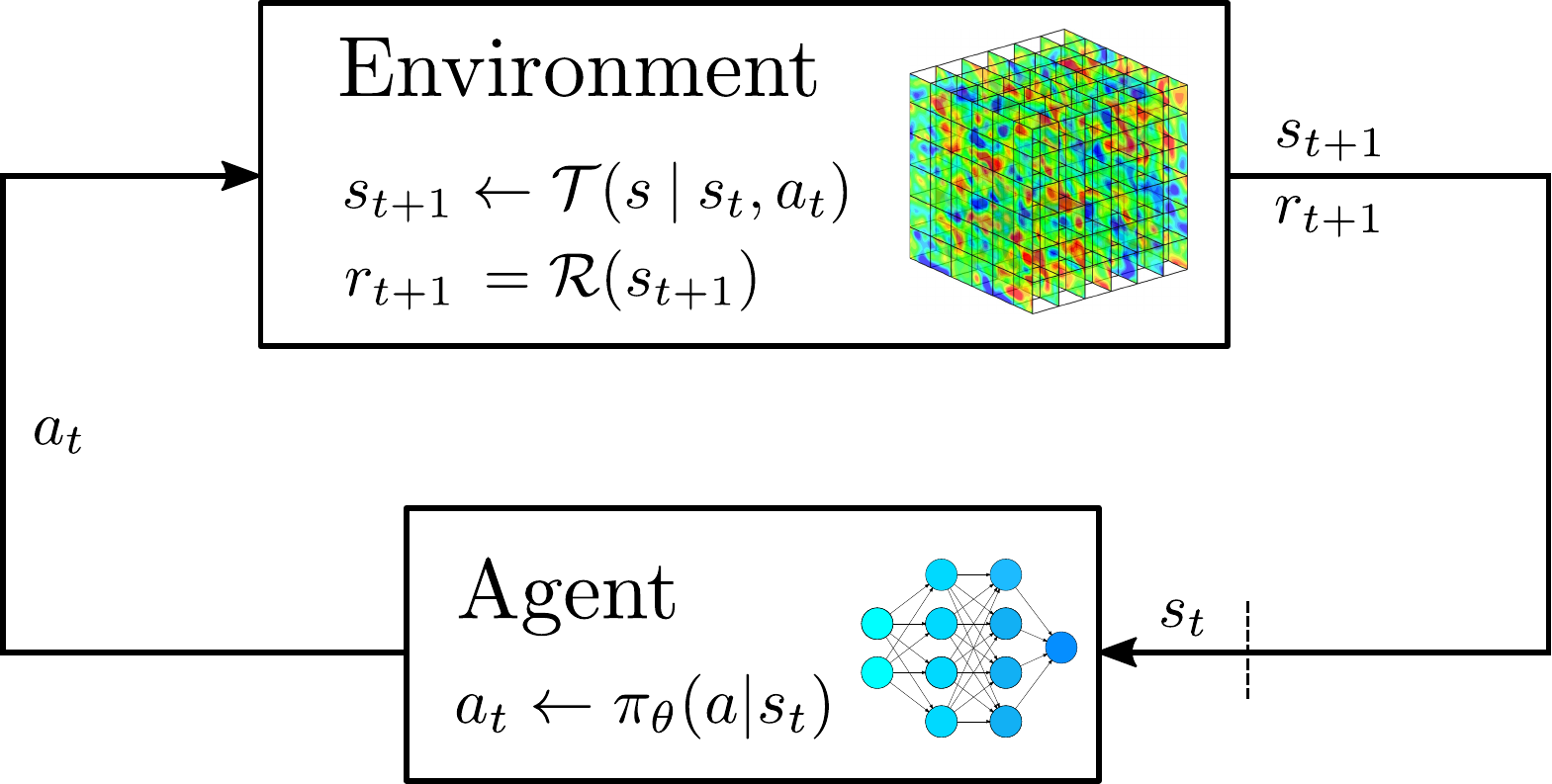}
  \caption{General outline of the Markov decision process (MDP). At step $t$, the environment is in state $s_t$. Following its policy $\pi_{\theta}(a\:|\:s_t)$, the agent performs an action $a_t$. In deep RL, the policy is a deep artificial neural network (ANN) with parameters $\theta$. The action causes the environment to transition from state $s_t$ to a new state $s_{t+1}$ that is prescribed by the environment's transition function $\mathcal{T}(s\:|\:a_t,s_t)$. Based on how desirable the new state is, the agent receives a reward, which is determined by the reward function $r_{t+1}=\mathcal{R}(s_{t+1})$.}
  \label{fig:MDP}
\end{figure}

The following section gives a brief outline of the general reinforcement learning (RL) paradigm.
However, this summary is by no means exhaustive and provides only the bare fundamentals required to motivate our software implementation, which is introduced in \secref{sec:software}.
For a more thorough discussion of RL, the reader is referred to \cite{sutton2018reinforcement}.
In the RL paradigm, an agent trains by interacting with an environment, as illustrated in \figref{fig:MDP}.
In each point in time $t$, the environment is in some state $s_t$, based on which the agent's (possibly parametrized) policy $\pi_{\theta}\left(a\:|\:s_t\right)$ prescribes which action $a_t$ the agent should perform.%
\footnote{In principal, the policy $\pi_{\theta}\left(a\:|\:s=s_t\right)$ is a random variable, which describes the conditional probability distribution of performing action $a$, given the state $s_t$. To keep the notation short, we will use $\pi_{\theta}\left(a\:|\:s_t\right) \equiv \pi_{\theta}\left(a\:|\:s=s_t\right)$ and refer the reader again to \cite{sutton2018reinforcement} for more details. The same holds for the transition function $\mathcal{T}\left(s_{t+1}\:|\:a_t,s_t\right)\equiv\mathcal{T}\left(s_{t+1}\:|\:a=a_t,s=s_t\right)$.}
This action causes the environment to change its state to a new state $s_{t+1}$, which is determined by the environment's transition function $\mathcal{T}(s_{t+1}\:|\:a_t,s_t)$.
The transition function thus encodes the dynamics of the environment, which could be for instance the spatial and temporal integration of the discretized Navier-Stokes equations.
Alongside the new state $s_{t+1}$, the agent receives a reward $r_{t+1}=\mathcal{R}(s_{t+1})$, which quantifies how favorable the transition is with respect to some performance metric.
The reward function $\mathcal{R}(s)$ is highly problem-specific and has to be designed by a domain expert.
Based on the new state $s_{t+1}$, the agent performs another action, until the environment reaches a final state $s_n$.
Eventually, such an episode results in a trajectory $\tau$ of states, actions and rewards:
\begin{equation}
  \tau = \left\{ \left(s_0,a_0\right),\left(s_1,a_1,r_1\right),\:......\;,\left(s_{n},a_{n},r_{n}\right)\right\}.
  \label{eq:trajectory}
\end{equation}
This problem formulation is typically framed as a Markov decision process (MDP), which again can be interpreted as a discrete-time control task.

The behavior of the agent is described by its policy $\pi_{\theta}$, which determines the action the agent performs for each of the possible states of the environment.
In deep RL, this policy is represented by an ANN with the weights $\theta$.
The key quantity to distinguish between favorable and unfavorable actions is the expected future return along a trajectory $\tau$ with $n$ steps
\begin{equation}
  R(\tau) = \sum_{t=1}^{n} \gamma^t r_{t} .
\end{equation}
Here, $\gamma \leq 1$ is the discount factor, which balances the importance of short-term and long-term rewards.
The overall goal of the RL algorithm is then to find the optimal policy, i.e. the set of optimal model parameters, which maximizes the expected return on all possible initial states.

The key purpose of an RL algorithm is to state an optimization task that allows optimization of the policy based on sampled interactions of the agent with the environment.
RL algorithms thus differ primarily in the way those interactions are sampled and how these interactions are used to update the policy.
Throughout this work, we employ the clipping version of proximal policy optmization (PPO) \cite{schulman2017proximal} as our RL algorithm of choice.
We highly recommend \cite{notter2021hierarchical} for a clear and concise summary of the PPO method.

\section{Software Architecture}\label{sec:software}
\begin{figure*}[htb]
  \centering
  \includegraphics[width=\linewidth]{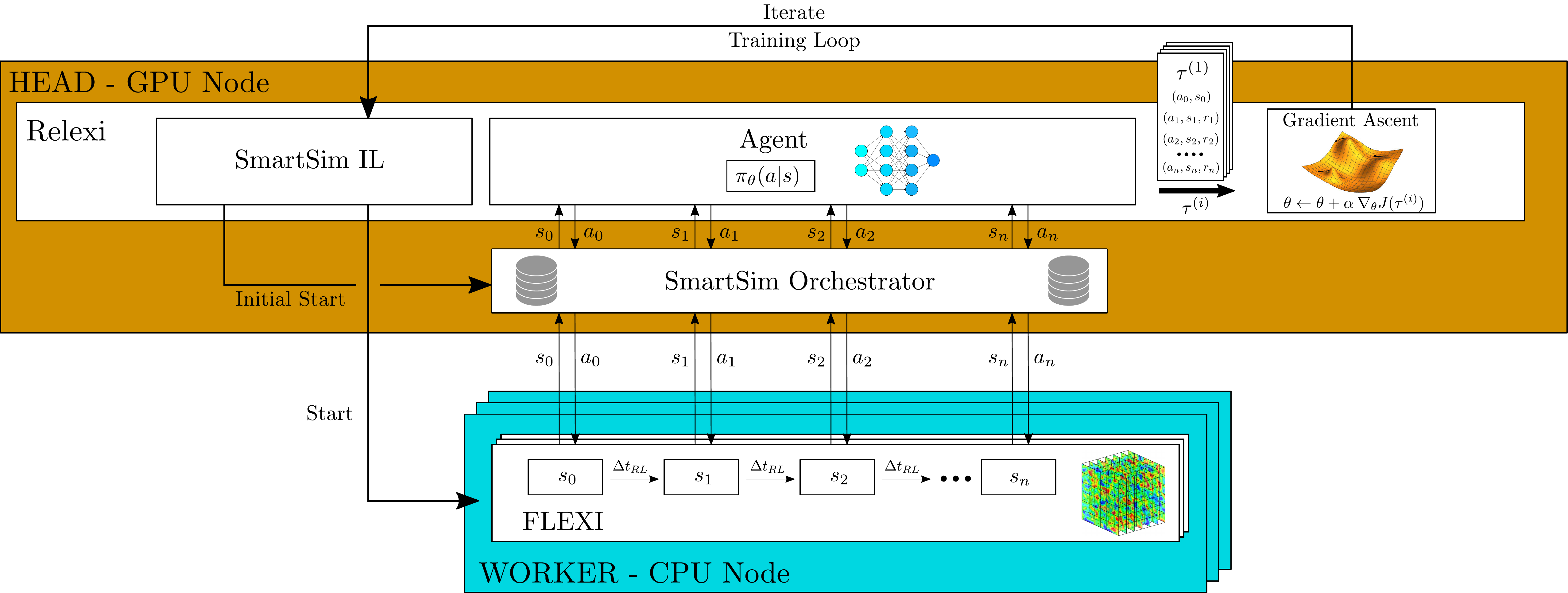}
  \caption{General architecture of Relexi. Before entering the training loop, the SmartSim IL is used to launch the SmartSim Orchestrator, which provides a database to exchange data between Relexi and the HPC workload. At the beginning of each training loop, the SmartSim IL starts a batch of FLEXI simulations and distributes them to the \textit{worker} nodes. The simulations are initialized on a randomly drawn initial state and are then evolved in time by the flow solver. For our application in turbulence modeling, FLEXI sends each time interval $\Delta t_{RL}$ its current state to the agent and receives a new action, i.e. a new set of parameters for the turbulence model. Since a syncronous RL algorithm is applied here, Relexi waits until all simulations have terminated. The collected episodes $\tau^{(i)}$ are then used by the gradient ascent algorithm to improve the model weights $\theta$. Here, $\nabla_{\theta} J(\tau^{(i)})$ is the gradient of the loss function, evaluated on the sampled experience. The training loop is then repeated with the new set of weights, i.e. the new policy, until the policy converges.}
  \label{fig:relexi}
\end{figure*}
In this work, we present a novel and modular RL framework named Relexi\footnote{Code available at: \href{https://github.com/flexi-framework/relexi}{https://github.com/flexi-framework/relexi}} that implements an RL training loop for applications in the field of scientific computing, as illustrated in \figref{fig:relexi}.
As discussed beforehand, coupling HPC codes with modern ML libraries is oftentimes tedious.
This holds especially for RL, where the interaction between the HPC application and the RL algorithm becomes much more intricate.
A considerable amount of these difficulties stem from the differences in the required hardware and the different preferences in terms of programming languages.
Many approaches in coupling HPC and ML tried to either rewrite basic ML capabilities in languages used for HPC or writing an HPC solver in Python from scratch.
Considering the enormous codebases on both sides and the enormous pace of progress in the ML community, this approach is, from our perspective, doomed to fail.
In addition, both families of methods require significant computational effort:
While the parameter update of the policy requires access to the collected trajectories to compute the gradient vectors through backpropagation, running CFD simulations mandates high parallel efficiency, in particular for problems of practical interest.
Thus, compromising the performance of any one component, e.g. through re-writing, is not advisable.
Relexi tries to bridge this gap differently by employing the SmartSim library to couple a state-of-the-art RL library with modern HPC solvers.
Special focus is put on the modularity of the framework, which allows the exchange of RL algorithms and simulation environments with only minimal changes of the underlying code. 
Relexi is designed with application on HPC system in mind, scaling applications up to thousands of cores on modern HPC systems.
Relexi comprises the following building blocks, which will be discussed in more detail in the following sections.
\begin{itemize}
  \item \textbf{SmartSim}: The SmartSim library \cite{partee2021using} fulfills two major tasks. Firstly, the library is used to start and manage the MPI-parallelized HPC workloads on the allocated hardware resources. Moreover, it implements the communication between the HPC workload and Relexi by deploying an in-memory database and providing communication clients for several programming languages.
  \item \textbf{FLEXI}: In this work, the flow solver FLEXI \cite{krais2021flexi} provides simulations of turbulent flow as a training environment for the agent. FLEXI is used here for illustrative purposes only and can be easily replaced by other simulation codes.
  \item \textbf{TensorFlow/TF-Agents}:  Relexi is build upon TensorFlow \cite{tensorflow2015-whitepaper}, which provides a framework for efficient ML workflows and allows for distributed training on multiple GPUs in parallel. In addition, its RL extension TF-Agents \cite{TFAgents} provides implementations of several RL algorithms and allows for custom environments that can be integrated in a straight-forward fashion.
\end{itemize}

\subsection{SmartSim}\label{subsec:smartsim}
The SmartSim library \cite{partee2021using} is a workflow library that simplifies the convergence of traditional HPC workloads and ML.
The library comprises two components: the SmartSim Infrastructure Library (IL) and SmartRedis.
The IL provides extensive functionalities to start, manage and distribute workloads in HPC environments as well as submitting workloads automatically as batch jobs to the job scheduler. 
In Relexi, we employ the IL to repeatedly start the MPI-parallelized simulation instances on demand for each training iteration inside of a single allocated batch job.
The simulations can be either started individually or multiple simulations can be lauched within a single call by using the multiple-program-multiple-data (MPMD) paradigm.
We also use another key aspect of the SmartSim IL, which is its ability to configure and launch an in-memory, Redis-based datastore referred to as the Orchestrator. 
This database serves as a data intermediary between the simulations and the main program.
For Relexi, we used a non-clustered Orchestrator, which is launched on the \textit{head} node.
SmartSim also supports the distribution of the Orchestrator across multiple nodes, but a single instance on the \textit{head} node was sufficient for our application.
The simulation communicates with the Orchestrator using the  SmartRedis clients (available in Python, Fortran, C, and C++) which can send and receive data from the orchestrator or trigger actions within the Orchestrator, i.e. running scripts or ML model inference previously loaded into the database.
In our case, during training, FLEXI sends its current flow state via the Fortran SmartRedis client to the Orchestrator.
In addition, a scalar flag is sent which indicates whether FLEXI has reached its final state and will terminate.
Relexi then uses the Python SmartRedis client to read this data from the Orchestrator.
Subsequently, the agent's actions are sent by Relexi to the database via its Python SmartRedis Client and are read by FLEXI via its Fortran client.
While SmartSim itself employs a Redis database by default, we used the multi-threaded fork of Redis called KeyDB, which provided significantly more performance for our application.

\subsection{FLEXI}\label{subsec:flexi}
The open-source flow solver FLEXI \cite{krais2021flexi} is based on the discontinuous Galerkin (DG) method, which can be seen as a hybrid of the finite element and the finite volume methods.
For the DG method, the computational domain is partitioned into individual elements.
In each element, the solution is represented by a polynomial basis of degree $N$ and the individual elements are then coupled by consistent surface fluxes at the element faces.
This results in a small communication stencil, which allows FLEXI to scale perfectly on hundreds of thousands of cores using a pure distributed-memory parallelization with MPI.
A detailed description of FLEXI can be found in \cite{krais2021flexi}.
The communication between FLEXI and the database is implemented by means of the SmartRedis clients provided by the SmartSim library.
To this end, SmartRedis is linked to FLEXI during compile time.
Implementing the data transfer in FLEXI is straight-forward and requires only a few lines of additional code.
Since FLEXI is parallelized with a distributed-memory approach, every rank contains only a chunk of the global flow state, which represents the environment's state $s_t$ in the RL formulation.
To communicate with the database, the flow state is thus first gathered across all ranks of the respective FLEXI instance before it is written to the database by the root rank.
Analogously, the predictions sent by Relexi are retrieved from the database by the root rank and are then scattered across the other ranks of the respective FLEXI instance.
It seems important to stress once again that FLEXI is used primarily for demonstration purposes and can be easily replaced by other simulation codes by using the SmartRedis Clients.

\subsection{Relexi}\label{subsec:relexi}
Relexi implements the main RL training loop by means of the TF-Agents \cite{TFAgents} library, as illustrated in \figref{fig:relexi}.
In TF-Agents, custom environments can be implemented by subclassing the provided environment class and implementing a few mandatory methods like initialization and time-stepping.
Since the TF-Agents library interacts with the environment via these pre-defined function interfaces, all intricacies of the coupling with the HPC workload are ``hidden'' inside of the environment class.
This has the advantage that all RL algorithms and tools provided by TF-Agents can interact natively with the custom environment and thus, with the HPC workload.
The hardware resources are distributed as follows.
Relexi itself is started as a single thread on a node termed \textit{head}.
The \textit{head} node executes all ML-specific work and thus should be equipped with a GPU.
Relexi also supports setups with multiple GPUs on the \textit{head} node by using the distribution strategies provided by TensorFlow.
The SmartSim Orchestrator is also started on the \textit{head} node at the beginning of Relexi.
During training, the HPC workload is repeatedly started with MPI on the available \textit{worker} nodes.
To ensure that each MPI rank is placed correctly on the available hardware and to avoid double occupancy, Relexi generates rankfiles on-the-fly based on the available hardware resources.
These rankfiles are then passed to MPI to ensure the correct placement of the MPI ranks.
 
After the simulations are launched, each simulation instance writes its initial states to the database.
Relexi reads these states, provides the respective actions based on the current policy and writes them back to the database.
Each FLEXI instance reads its prescribed actions from the database and proceeds with its simulation to obtain the next state. %
In the meantime, Relexi polls until the new state becomes available and reads it from the database.
With the new state, the agent can compute the reward and get the new actions from the current policy.
These steps are repeated until the required amount of experience is sampled, on which the policy then can be trained.
This algorithmic outline of Relexi is also summarized in \algoref{alg:relexi}.%
\footnote{For the sake of clarity, \algoref{alg:relexi} only gives the algorithm for a single environment. To generalize it to multiple environments in parallel, each of the lines 7 to 11 is simply executed for each individual environment.}

\begin{algorithm}
  \caption{Relexi}
  \begin{algorithmic}[1]
    \State Initialization
    \State Launch SmartSim Orchestrator
    \For{$i = 1,i_{max}$}  \Comment{Run $i_{max}$ iterations}
       \State Start FLEXI instances
       \State Read $s_{0}$
       \For{$t=1,n$}  \Comment{Run simulation for $n$ steps}
         \State $a_t \leftarrow \pi_{\theta}(a\:|\:s_t)$
         \State Write $a_t$
         \State Polling for $s_{t+1} $\Comment{Here, the HPC solver runs}
         \State Read $s_{t+1}$
         \State $r_{t+1}\leftarrow \mathcal{R}(s_{t+1})$
       \EndFor
       \State $\tau \leftarrow \{ (s_0,a_0), (s_1,a_1,r_1), ..)\}$
       \For{$j=1,n_{epochs}$}  \Comment{Train ANN for $n_{epochs}$}
         \State $\theta \leftarrow \theta + \alpha \, \nabla_{\theta} J(\tau)$
       \EndFor
    \EndFor
    \State Shutdown SmartSim Orchestrator
  \end{algorithmic}
  \label{alg:relexi}
\end{algorithm}

A potential bottleneck we identified is the overhead introduced by repeatedly starting hundreds of parallel environments with thousands of MPI ranks.
For some configurations, the time required for starting the simulations exceeded the actual simulation time.
To tackle this issue, we implemented two major improvements.
First, we employed the multiple-program-multiple-data (MPMD) functionality provided by OpenMPI's implementation of MPI, which is also supported by SmartSim.
With MPMD, all simulations can be started with individual commandline arguments within a single call of MPI.
Secondly, we implemented a functionality to copy all files required by the simulation, e.g. parameter files and restart files, to local drives located in the random access memory (RAM) of each node.
This reduced the access times compared to using a parallel file system like Lustre significantly.
With these improvments in place, the performance penality of launching large amounts of environments became negligible.

\section{Hardware Configuration}\label{sec:hardware}

    All benchmarks and experiments are performed on the HPE Apollo 9000 supercomputer (Hawk) and the Hawk-AI extension, a HPE Apollo 6500 Gen10 Plus at the High-Performance Computing Center Stuttgart (HLRS).
    In the following, the hardware of both systems is given in detail.

    \subsection{Hawk -- HPE Apollo 9000}

        Hawk consists of 5,632 dual socket nodes with 256 GiB of main memory each.
        Each node is equipped with two 64-core AMD EPYC 7742 (Rome) processors with a base frequency of 2.25 GHz.
        The nodes are connected via an enhanced 9D-hypercube.
        For the node to node interconnect, the high-performance interconnect InfiniBand HDR200 is used.
        This leads to a homogeneous massively parallel system with 720,896 compute cores and approximately 1.37 TiB of main memory.
        Hawk has a theoretical peak-performance of 25.1 Pflop/s.
        The system reached 19.334 Pflop/s in the LINPACK-benchmark and 334.65 TFlop/s in the HPCG benchmark.
        Hawk is connected to a Lustre file system with a capacity of about 25 PiB.

    \subsection{Hawk-AI -- HPE Apollo 6500 Gen10 Plus}
        The Hawk-AI extension consists of 24 dual socket nodes, which are each equipped with two 64-core AMD EPYC 7702 processors, 1 TiB of main memory, a local hard disk of 15 TiB and eight Nvidia A100 GPUs.
        20 Nodes are equipped with Nvidia A100 with 40 GiB of memory and four nodes are equipped with Nvidia A100 with 80 GiB of memory.
        The nodes are also connected via the InfiniBand HDR200 interconnect.
        The Hawk-AI extension is connected directly to the enhanced 9D-hypercube of Hawk and to the same Lustre filesystem.%

\section{Application to Turbulence Modeling}\label{sec:numerics}

\subsection{Turbulence Modeling}
Turbulent flows are notoriously hard to resolve accurately, since turbulence is a multiscale phenomenon.
The resolution requirements for numerical simulations to resolve the wide range of active length scales are usually intractable.
Instead, reduced order descriptions of turbulence can be derived, which only resolve the large energy-containing flow scales, while employing a turbulence model to account for the influence of the non-resolved fine scales.
From a mathematical standpoint, this is equivalent to applying a low-pass filter to the flow field.
This approach is commonly referred to as large eddy simulation (LES).
A myriad of different LES turbulence models have been proposed in literature.
However, such models typically contain empirical parameters, which have to be tuned to the employed numerical discretization and the specific test case.
As a consequence, no universal or overall \emph{best} model has been found to this date.

The most important property of the LES model is to mimic the energy drain from the large to the small flow scales.
Based on this reasoning, a popular modeling strategy is to approximate the unresolved scales by introducing an additional turbulent viscosity $\nu_t$ that is added to the physical viscosity.
The common model by Smagorinsky \cite{smagorinsky1963general} computes this turbulent viscosity as
\begin{equation}
  \nu_t= \left(C_s \Delta\right)^2 \sqrt{2\:\overline{S}_{ij}\overline{S}_{ij}} \qquad \text{with} \quad \overline{S}_{ij} = \frac{\partial \overline{u}_i}{\partial x_j}.
  \label{eq:smagorinsky}
\end{equation}
Here, $\overline{S}_{ij}$ is the rate-of-strain tensor of the coarse-scale velocity field $\overline{u}_i$ with respect to the coordinates $x_j$ with $i,j=1,2,3$.
The filter width $\Delta$ is a measure of the employed spatial resolution and $C_s$ is a model coefficient, which has to be tuned for each specific test case.
Another common modeling approach is the implicit modeling paradigm, which assumes that the numerical dissipation error introduced by the numerical scheme acts as an implicit LES model. The implicit LES model can obviously be seen as a special case of Smagorinsky's model with $C_s=0$.

With our RL framework, we strive to improve the performance of Smagorinsky's model given in \eqref{eq:smagorinsky} by employing an RL algorithm to tune the model coefficient $C_s$ dynamically in space and time during the simulation.
In this regard, we show that RL represents a promising approach to enhance current simulation workloads and that our novel RL framework is capable of handling computationally intensive simulation environments on modern HPC architectures.

\subsection{Stating the Reinforcement Learning Task}
As test case, i.e. as training environment, we perform LES of homogeneous isotropic turbulence (HIT) at a Reynolds number of $Re_{\lambda}\approx 200$ with respect to the Taylor microscale.
The cubic domain of side length $2\pi$ is equipped with periodic boundary conditions and is discretized by a Cartesian mesh with the resolutions given in \tabref{tab:testcases}.
This test case of \emph{turbulence in a box} can be seen as the building block of turbulence and describes freely decaying turbulence in the absence of boundaries.
In order to obtain a quasi-static solution, an isotropic linear forcing is employed as proposed by \cite{lundgren2003linearly,de2015anisotropic} to balance the dissipation of the turbulence model.
This results in a quasi-stationary distribution of the turbulent kinetic energy in the system, which is mainly characterized by the energy drain from the large to the small scales.
Since this energy cascade is a fundamental property of turbulence that the LES should reproduce, we define our optimization target for the RL task in terms of this energy spectrum.
The reward for the RL algorithm is thus computed based on the error of the instantaneous energy spectrum of the LES $E_{LES}(k)$ compared to the mean energy distribution of the underlying ground truth solution $E_{DNS}(k)$, which was obtained beforehand from a high-fidelity simulation.
For this, we use the mean relative error across the wavenumbers $k$ up to $k_{max}$, which can be computed as
\begin{equation}
  \ell = \underset{k}{\mathrm{mean}}\left[\left(\frac{E_{DNS}(k)-E_{LES}(k)}{E_{DNS}\left(k\right)}\right)^2\right], \quad k\in\left[1,k_{max}\right].
  \label{eq:reward_error}
\end{equation}
To ensure that the reward is normalized to $r_t\in[-1,1]$, the reward is eventually computed as
\begin{equation}
  r_t = 2 e^{\left(\ell / \alpha \right)}-1,
  \label{eq:reward}
\end{equation}
with $\alpha$ as a scaling parameter.

With the reward function in place, the optimization task for the RL algorithm is framed as follows.
The state of the environment observed by the agent is the current coarse-scale velocity field $\overline{u}_i$.
The agent predicts as action a single $C_s$ coefficient for each element in the computational mesh solely based on the local flow state in the respective element.
The environment's state is then evolved with the flow solver for some time $\Delta t_{RL}$ before requesting new predictions for the elementwise $C_s$.
The time inteval $\Delta t_{RL}$ is generally chosen much larger than the computational timestep of the simulation.
The reward for the agent is computed based on the differences in energy distribution compared to the ground truth, as given in \eqref{eq:reward}.
This loop is repeated until some final time $t_{end}$ is reached.

\subsection{Computational Setup}
In this work, we perform LES at different resolutions, which are listed in \tabref{tab:testcases}.
All simulations are run up to $t_{end}=5$ and actions are performed with in a time interval of $\Delta t_{RL}=0.1$, which corresponds to 50 predictions per simulation.
The initial state for each simulation run is drawn randomly from a set of flow states that are computed by filtering different realizations of the high-fidelity solution.
A single initial state is kept hidden to evaluate the model performance on unseen test data.
Based on these setups, two different approaches can be followed to reduce the required computational time for the training by using parallel HPC resources.
First, the number of MPI ranks per environment can be increased to reduce the simulation time of the environments by exploiting the strong scaling capabilities of FLEXI.
Second, the number of simulated environments per training update can be increased, which can be seen as a weak scaling approach for the full framework.
Since an increased number of sampled episodes might result in a better estimate of the gradient for the optimizer, this approach can reduce the number of iterations required for convergence and thus the necessary training time.
Since both perspectives highlight crucial scaling properties of the underlying framework, both will be investigated in \secref{sec:scaling}.
Here, we employ up to 16 cores per FLEXI environment and up to 1024 parallel environments per training iteration using a maximum of 2,048 cores.
\begin{table}
  \centering
  \begin{tabular}{lccccc}
    \hline    
          & N & \#Elems & \#DOF  & $k_{max}$ & $\alpha$ \\
    \hline    
    24 DOF & 5 &   $4^3$ & 13,824 &         9 &   0.4  \\
    32 DOF & 7 &   $4^3$ & 32,768 &        12 &   0.2  \\
    \hline    
  \end{tabular}
  \caption{Investigated configurations for the LES of the HIT test case. The runs are named by the number of degrees of freedom (DOF) per spatial direction, which result from the given polynomial degree $N$ and the number of elements \#Elems. The total number of DOF can be computed by $\text{\#DOF} = \text{\#Elems}\:(N+1)^3$. The hyperparameters $k_{max}$ and $\alpha$ refer to the maximum wavenumber and the scaling factor of the reward function, respectively.}%
  \label{tab:testcases}
\end{table}

As RL algorithm, we use the clipping variant of the proximal policy optimization (PPO) \cite{schulman2017proximal} algorithm as already discussed in \secref{sec:rl}.
This algorithm is synchronous, which means that the algorithm performs the steps of sampling experience and updating the policy in a sequential fashion.
This means that first, FLEXI simulations are performed based on the current policy and thereafter, the policy is updated based on the collected experience to maximize the reward on future runs.
For all experiments we used a discount factor of $\gamma=0.995$ and a learning rate of $10^{-4}$ with the Adam optimizer \cite{kingma2014adam} to train the policy for 5 epochs per iteration.
For the PPO algorithm, we used a clipping parameter of 0.2 and set the entropy coefficient to zero.
The employed policy ANN comprises around 3,300 parameters with its architecture given in \tabref{tab:ann}.

\begin{table}
  \centering
  \begin{tabular}{lrrrr}
    \hline    
    Layer   & Kernel & Filters & Padding  & Dimension \\
    \hline     
    Input   &        &         &          &  $6\times 6\times 6\times 3$ \\
    Conv3D  &      3 &       8 &     zero &  $6\times 6\times 6\times 8$ \\
    Conv3D  &      3 &       8 &     none &  $4\times 4\times 4\times 8$ \\
    Conv3D  &      3 &       4 &     none &  $2\times 2\times 2\times 4$ \\
    Conv3D  &      2 &       1 &     none &  $1\times 1\times 1\times 1$ \\
    Scale   &        &         &          &  $                        1$ \\
    \hline    
  \end{tabular}
  \caption{Architecture of the policy ANN for $N=5$ with the dimensions of each layer's output. The model's input dimensions follow from the $(N+1)^3$ solution points in each element times the three velocity components $\overline{u}_i$, which are used as input features. The ANN is build from three-dimensional convolutional layers with a specific kernel size and number of filters, each. The first layer uses zero-padding, while the rear layers employ no padding to convolve the high-dimensional input to a single scalar. All layers except the last convolutional layer employ the rectified linear unit (ReLU) as activation function. The final scaling layer uses the operation $y=\frac{1}{2}\sigma_s(x)$ to scale the input to the interval $[0,0.5]$ with $\sigma_s(x)$ denoting the sigmoid activation function.}
  \label{tab:ann}
\end{table}

\section{Results}\label{sec:results}

\subsection{Scaling}\label{sec:scaling}

\begin{figure*}[htb]
  \centering
  \includegraphics[width=\textwidth]{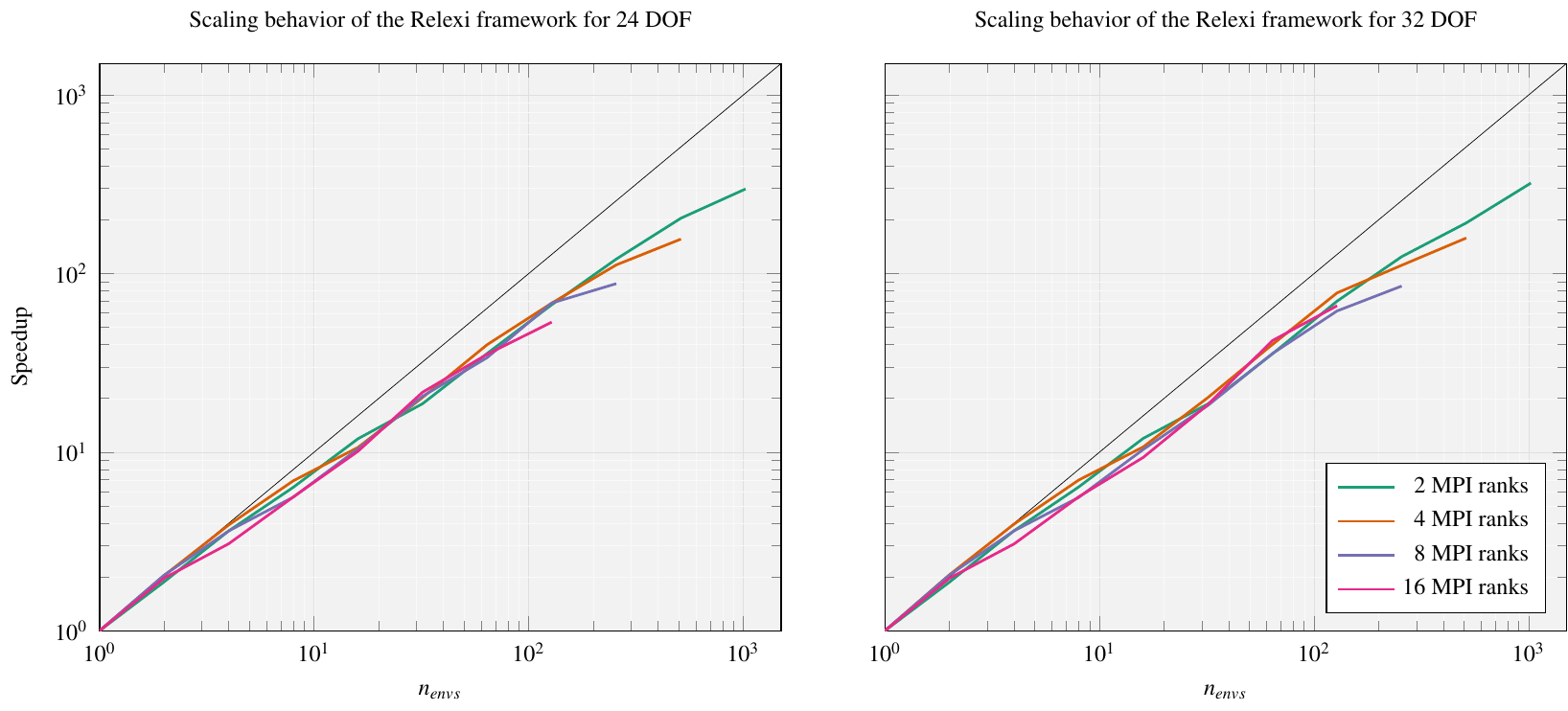}
  \caption{Scaling behavior of the Relexi framework on up to 16 Hawk compute nodes (2048 MPI ranks) and one Hawk-AI node for the HIT test case with 24 DOF and 32 DOF for 2, 4, 8 and 16 MPI ranks per FLEXI instance. The black line indicates perfect scaling.}
  \label{fig:scaling_hit}
\end{figure*}

\begin{figure*}[htb]
  \centering
  \includegraphics[width=\textwidth]{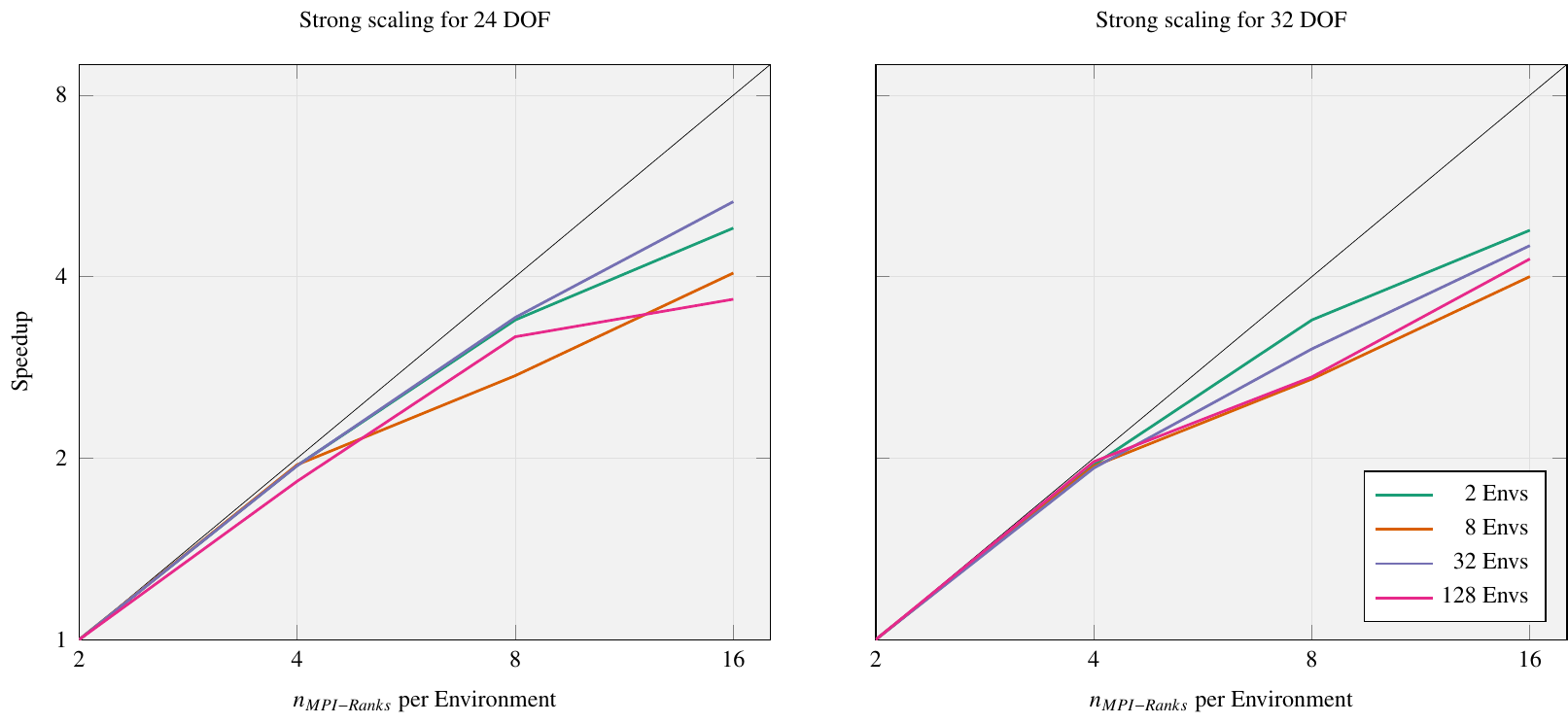}
  \caption{Strong scaling of FLEXI within the Relexi framework for the HIT test case with 24 DOF and 32 DOF for 2, 4, 8 and 16 MPI ranks per FLEXI instance. For clarity, only the runs for 2, 8, 32 and 128 parallel FLEXI environments are shown. The black line indicates perfect scaling.}
  \label{fig:strong_scaling_hit}
\end{figure*}

To analyze the scaling behaviour of our framework, we use the HIT test case for the two different configurations listed in \tabref{tab:testcases}.
For the benchmarks, we used a single Hawk-AI node of the HPE Apollo 6500 Gen10 Plus as \textit{head} node and up to 16 Hawk compute nodes of the HPE Apollo 9000 as \textit{worker} nodes in a single batch job.
This results in up to 2048 compute cores used for generating training data.
In the scaling analysis, we investigate two types of scaling.
First, we examine how the framework scales if the number of parallel environments is increased, while keeping the amount of MPI ranks per environment constant.
This corresponds to a weak scaling approach, since the parallel load and the available resources are increased simultaneously, while keeping the load per rank constant.
Since the used resources, and thus also the simulation time, for a FLEXI simulation should theoretically stay unchanged, eventually emerging losses in parallel speedup can be attributed solely to the communication overhead, the limited throughput of the database, the cost of the data managment in Relexi, the increasing amount of policy evaluations and the start as well as the termination of the simulations.
The measured executation time includes launching the simulations and running the simulation with the policy until all simulations terminated.
As a performance metric, we use the ``Speedup'', which is computed as the quotient of the time needed for sampling the $n_{envs}$ parallel environments and the time that would be needed to run $n_{envs}$ environments sequentially, i.e. the speedup of running the environments in parallel instead of sequentially.
For each configuration, we ran Relexi in two separate jobs for 6 iterations each to account for fluctuations in hardware and communication performance and computed the mean of the 12 measurements.
The scaling is performed with either 2, 4, 8 or 16 MPI ranks for each FLEXI run.
Further, we start with two parallel FLEXI instances and double them until the 16 Hawk compute nodes are fully occupied.
It seems important to note that the results are not compared directly to the performance of standalone FLEXI simulations, since the performance differences in running a single FLEXI instance within Relexi and running it standalone were negligible.

The results in \figref{fig:scaling_hit} demonstrate that the framework can scale efficiently up to a thousand parallel environments on thousands of cores.
Two major trends can be identified in the scaling results.
As mentioned before, the observed decrease in parallel efficiency when running more environments in parallel can be attributed mainly to the sequential work done by Relexi.
The framework should scale better if the FLEXI instance gets more time-consuming, since then, this sequential work becomes less relevant and the perfect scaling abilities of FLEXI can be recovered.
In contrast, if the necessary time to compute the FLEXI simulation decreases, i.e. FLEXI gets more ranks, the sequential work of Relexi becomes more dominant, which decreases the scaling efficiency.
This causes the runs with fewer ranks per FLEXI instance to scale better than the runs using more ranks.
The second interesting behavior is that the decrease in performance when switching from a single to two parallel environments is most pronounced for the FLEXI environment with only two cores, which is counterintuitive. 
We attribute this to the limited memory bandwith and the hierarchical architecture of the processors used.%
\footnote{The used EPYC CPUs comprise several dies, which contain 8 cores each. All cores on a single die share the available memory bandwidth.}
If a single FLEXI instance with two ranks is spawned on a compute node, this instance gets all available memory bandwidth.
If however, a second FLEXI is spawned, both instances have to share the available bandwidth, which slows down the simulation as well as the interaction with the database.
This effect vanishes with an increasing amount of used cores.
The observed loss of parallel efficiency especially for the last data points, which correspond to using all 2,048 available cores, can in parts be attributed to single simulations running significant slower than the average.
These outliers can probably be attributed to fluctuations in the load of the interconnect. %
This issue is subject of current investigations.

For the strong scaling, we examine the configurations with 2, 8, 32 and 128 parallel FLEXI instances.
The results of the strong scaling given in \figref{fig:strong_scaling_hit} match the general behavior observed in the weak scaling benchmark. %
In cases where the computational cost of the HPC workload is dominant, i.e. if only few MPI ranks are used per environment, the optimal scaling of FLEXI can be recovered.
If the amount of MPI ranks is increased, the time spent for the simulation decreases and the work done by Relexi becomes the limiting factor.
This causes the parallel efficiency to decrease for low simulation loads per core.
It seems important to note that for both cases, using 16 MPI runs per simulation falls quite below the optimal load per core for FLEXI.
For the more realistic cases of up to 8 ranks per FLEXI, most of the FLEXI performance can be recovered.

\subsection{Turbulence Modeling}
\begin{figure*}[htb]
  \centering
  \includegraphics[width=\textwidth]{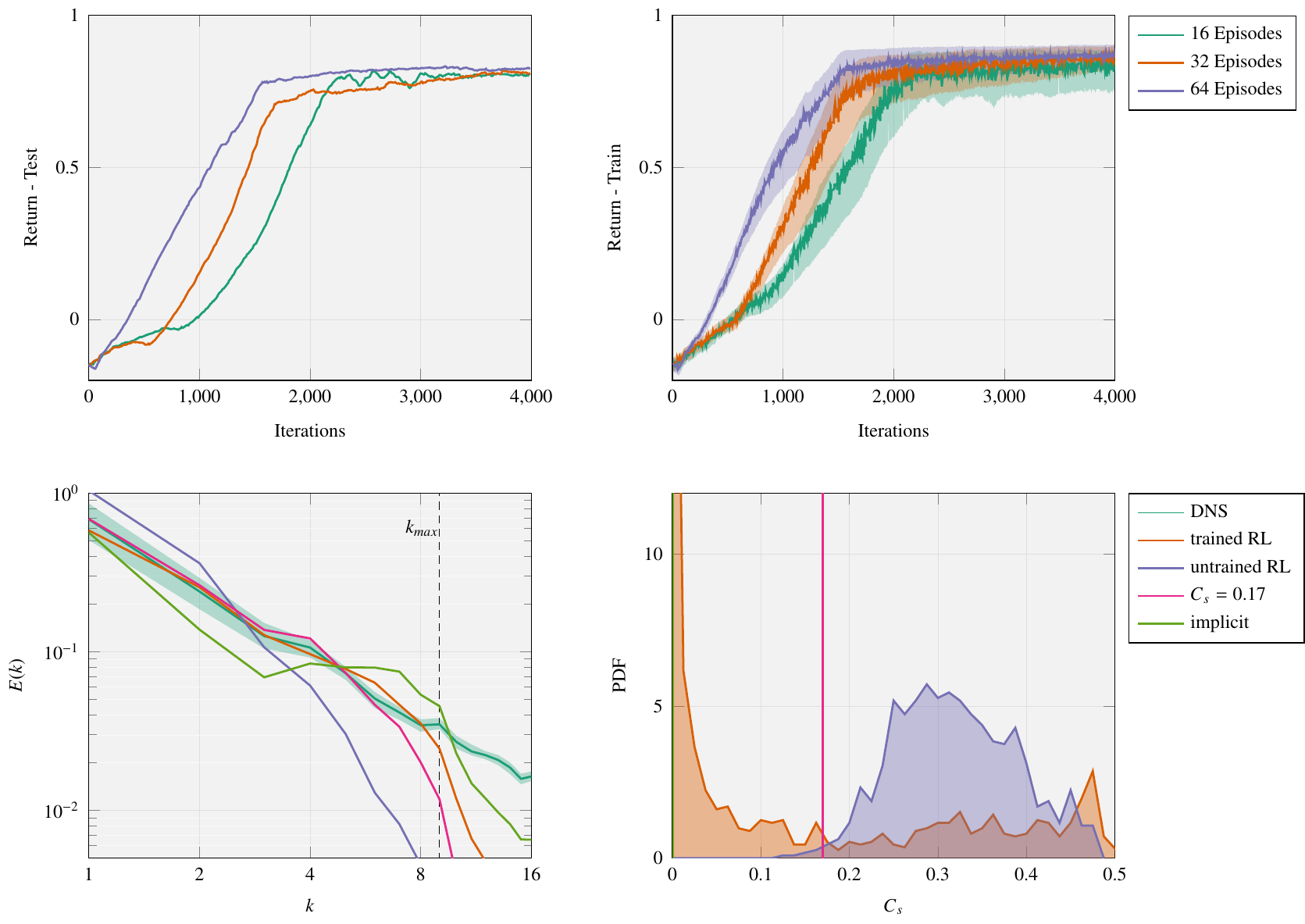}
  \caption{Training results of the 24 DOF configuration with 16, 32 and 64 sampled episodes per training update. Top left: Collected return averaged over all training runs and normalized with the maximum achievable return. The minimum and maximum return in each iteration is indicated by the shaded area. Top right: Normalized return on the unseen testing state, which was evaluated every 10 iterations. Bottom left: Spectra at the end of the simulation $t_{end}$ on the unseen test state after 4,000 training iterations for the configuration running 64 parallel environments. The resulting spectra of Smagorisky's model and the implicit model are shown for comparison. The shaded area around the time-averaged DNS spectrum indicates the maximum and minimum observed amplitudes during the DNS. Bottom left: Distribution of the model's predictions for $C_s$ during the simulation on the test state.}
  \label{fig:results_rl_24DOF}
\end{figure*}

To investigate the performance of the RL-based turbulence model against established analytical closure models, we trained the agent on the 24 DOF configuration for 4,000 Iterations, which corresponds to 20,000 gradient ascent steps for the policy overall.
To quantify the benefits of gathering more episodes per policy update, the training was carried out using 8 MPI ranks per environment and 16, 32 and 64 parallel FLEXI instances, respectively.
For the 16 and 64 runs configurations, the overall training required 20 and 30 hours, respectively.
Sampling the trajectories took 15 and 18 seconds per iteration, respectively, while updating the policy on a single GPU took 0.5 and 2 seconds, respectively.
The results in \figref{fig:results_rl_24DOF} indicate that increasing the number of parallel simulations does indeed improve the training performance in several ways.
Firstly, the training runs with more episodes converge faster, i.e. need less training iterations to achieve a given episode return. 
Moreover, the collected return of the 16 episode configuration increases less consistently after around 2,000 iterations than the training configurations with more parallel episodes.
This confirms our prior assumption that the gradient estimator becomes more reliable if more episodes are sampled, which again leads to more efficient training updates and thus to faster convergence.
In the same vein, the training runs using more episodes tend to converge to a higher total return.
This notion holds especially for the return collected during training, where the return is computed on a variety of different initial states.
While the return on the unseen initial state for testing gives similar results, the reward of the 16 episodes configuration fluctuates between 2,000 and 3,000 iterations and temporarily exceeds the return of the 32 episodes configuration before converging to a stable return.
This behavior probably stems from the high variance of the 16 episodes training which causes the less consistent improvement on the unseen test state.
The results in \figref{fig:results_rl_24DOF} also clearly demonstrate that for our application, the RL-based model outperforms both Smagorinsky's model and the implicit modeling strategy with regard to the energy spectrum.
Especially surprising is that the RL-enhanced model agrees with the high-fidelity data almost up to $k_{max}$ with deviations only at $k=6$.
This indicates that our novel data-driven model does not only replicate the vital flow physics related to the energy cascade, but also pushes the resolution limits of the underlying discretization.
The predictions of the initial untrained model appear to be almost normally distributed, as is expected from the initialization of the policy ANN.
Interestingly, the trained model modifies this distribution heavily by predicting tiny $C_s$ parameters for the vast majority of the flow field while increasing the parameter only in selective elements. 
The model also takes advantage of the entire admissable range of $C_s\in[0,0.5]$.
The reported results demonstrate that the application of the RL paradigm to turbulence modeling (and control tasks in CFD in general) can contribute to major advancements of the state-of-the-art given that the necessary resources can be used efficiently on massively parallel systems.

\section{Conclusions}\label{sec:conclusion}
Supervised learning is generally suitable for learning in situations when input--output pairs can be defined a priori.
In contrast, in reinforcement learning, training data is gathered in an online process, in which the policy is sampled for the current state of the dynamical system.
This makes RL more suitable for learning optimal behaviors in dynamical systems, e.g. those described by the equations of fluid dynamics, where an a priori definition of an admissible and complete training data space is illusive and would, at best, be cumbersome.
Flow control problems thus lend themselves naturally to an RL approach, however, other modeling tasks can be expressed in this context as well.
Here, we have chosen to interpret the task of finding an optimal eddy viscosity in space and time as a control problem, and have the RL-trained agent predict a strategy.
Our results show that this approach outperforms existing, established models and indeed returns a near optimal behavior in the chosen reward norm.
This highlights the potential of combining CFD and RL into an inclusive optimization framework.

However, before we can explore or leverage this potential, we need to enable training and deploying such algorithms at scale, recognizing that they pose different challenges for software development and HPC.
In RL, trajectory data consisting of both actions and states needs to be gathered along the solution evolution during training and made available to the gradient update.
The RL agent and the CFD scheme thus need to be coupled during training, and they need to exchange large datasets continuously. In addition, the current policy must be explored efficiently, meaning that many parallel runs of the environments are required.
Thus, in this work, we propose Relexi as a novel framework for coupling RL algorithms with essentially generic solvers for partial differential equations on heterogeneous hardware.
The framework is made up of three components:
The PDE solver of choice, in our case, the LES solver FLEXI, TensorFlow for the ML definition and training, and the SmartSim library, which handles job allocation and management and provides an in-memory database for communication and intermediate storage of solution trajectories and policy commands.
The PDE solver and TensorFlow can run independently on their chosen hardware, thus exploiting their full parallel potential.
We show two methods for improving the overall time-to-solution of the RL problem with Relexi:
First of all, the weak scaling across the gradient estimator, that is increasing the amount of runs of the environment for a given policy.
More parallel runs explore the current policy more efficiently and thus allow an overall better gradient update and thus quicker convergence.
Our scaling results here indicate that the framework scales up to hundreds of parallel environments and thousands of compute cores with very good performance.
For the second approach, we scaled the individual simulation runs across more MPI ranks, i.e. we exploit the strong scaling characteristics of the standalone PDE solver.
Here, we recover the expected behavior up until the individual core load becomes too small.
In combination, these results demonstrate that the Relexi framework is capable of efficient training on HPC systems at scale and can enable RL-methods for CFD for complex flow cases.
Further work will focused on applying Relexi to more complex cases and to other combinations of RL and PDE solvers.

\section*{Acknowledgment} %
The research presented in this paper was funded by Deutsche Forschungsgemeinschaft (DFG, German Research Foundation) under Germany's Excellence Strategy - EXC 2075 - 390740016.
This work was possible due to kind funding by the Ministerium für Wissenschaft, Forschung und Kunst Baden-Württemberg via the project SiVeGCS-MWK.
The authors gratefully acknowledge the support and the computing time on ``Hawk'' provided by the HLRS through the project ``hpcdg'' and the support by the Stuttgart Center for Simulation Science (SimTech).
Moreover, the authors gratefully acknowledge the SmartSim developers for their support.

\bibliographystyle{elsarticle-num}
\bibliography{draft}

\end{document}